\documentclass{article}
\usepackage{microtype}
\usepackage{graphicx}
\usepackage{booktabs}
\usepackage{hyperref}
\usepackage[algo2e,noend,boxruled]{algorithm2e}
\usepackage{subcaption}
\usepackage{caption}
\usepackage[accepted]{icml2025} 
\usepackage{amsmath}
\usepackage{amssymb}
\usepackage{mathtools}
\usepackage{amsthm}
\theoremstyle{plain}

\theoremstyle{definition}

\theoremstyle{remark}

\input{helpers}
\icmltitlerunning{Faster Language Models with Better Multi-Token Prediction Using Tensor Decomposition}
\begin{document}
\renewcommand{\ICML@appearing}{Copyright 2025 by the author(s).}
\twocolumn[

\icmltitle{Faster Language Models with Better Multi-Token Prediction\\Using Tensor Decomposition}

\icmlsetsymbol{equal}{*}

\begin{icmlauthorlist}
\icmlauthor{Artem Basharin}{airi,sk}
\icmlauthor{Andrei Chertkov}{airi,sk}
\icmlauthor{Ivan Oseledets}{airi,sk}
\end{icmlauthorlist}

\icmlaffiliation{sk}{Skolkovo Institute of Science and Technology}
\icmlaffiliation{airi}{Artificial Intelligence Research Institute (AIRI)}

\icmlcorrespondingauthor{Artem Basharin}{basharin@airi.net}

\icmlkeywords{LLM, Low rank decomposition}

\vskip 0.3in
]

\printAffiliationsAndNotice{} 
\begin{abstract}
We propose a new model for multi-token prediction in transformers, aiming to enhance sampling efficiency without compromising accuracy. Motivated by recent work that predicts the probabilities of subsequent tokens using multiple heads, we connect this approach to rank-$1$ canonical tensor decomposition. By generalizing it to a rank-$r$ canonical probability decomposition, we develop an improved model that predicts multiple tokens simultaneously. This model can also be interpreted as a mixture of experts, allowing us to leverage successful techniques from that domain for efficient and robust training. Importantly, the overall overhead for training and sampling remains low. Our method demonstrates significant improvements in inference speed for both text and code generation tasks, proving particularly beneficial within the self-speculative decoding paradigm. It maintains its effectiveness across various model sizes and training paradigms, highlighting its robustness and scalability.
\end{abstract}
\section{Introduction}
    Autoregressive transformer models~\citep{vaswani2017attention} have become a cornerstone in natural language processing tasks due to their ability to model complex sequential data.
However, one significant limitation of these models is the inefficiency in sampling during inference, as they generate tokens one at a time, leading to increased latency in practical applications~\citep{fournier2023practical, fields2024survey}.
Accelerating the inference process without compromising the model's performance is thus a critical challenge.

Recent efforts have explored multi-token prediction to address this inefficiency. A simple yet effective approach~\citep{gloeckle2024better} involves using multiple heads to predict the next $n$ tokens simultaneously.
This method approximates the joint probability of the next $n$ tokens by assuming conditional independence given the previous context.
Mathematically, given a sequence $(x_1, x_2, \ldots, x_t)$ this approximation can be expressed as:
\begin{equation}\label{llmtelora:rank-1}
P_{\theta}(x_{t+n:t+1} \vert x_{t:1})
\approx
\prod_{s=1}^{n}
    P_{\theta}^{(s)}(x_{t+s} \vert x_{t:1}).
\end{equation}
This equation represents a rank-$1$ tensor approximation of the joint probability distribution, effectively treating future tokens as independent of each other given the past tokens.
While this assumption simplifies computation and can be combined with speculative decoding~\citep{leviathan2023fast} to accept some of the predicted tokens, it remains a crude approximation that may limit token acceptance rates due to its disregard for token interdependencies.

To improve upon this, we propose a more accurate approximation of the joint distribution by introducing a sum over multiple rank-$1$ terms. Specifically, we generalize the approximation to a rank-$r$ canonical decomposition~\citep{harshman1970foundations, kolda2009tensor, cichocki2016tensor}:
\begin{equation}\label{llmtelora:cp}
P_{\theta}
    (x_{t+n:t+1} \vert x_{t:1
})
\approx
\sum_{\alpha=1}^{r}
    w_{\alpha} \prod_{s=1}^{n}
        P_{\theta}^{(s)}(x_{t+s} \vert x_{t:1}, \alpha),
\end{equation}
where learnable weights $w_{\alpha}$ ($\alpha=1,2, \ldots,r$) are satisfying
\begin{equation}\label{llmtelora:tens_norm}
w_{\alpha} \geq 0,
\quad
\sum_{\alpha=1}^r w_{\alpha} = 1.
\end{equation}

The proposed formulation in~\eqref{llmtelora:cp} accounts for dependencies among future tokens by effectively considering a mixture of expert models, each capturing different aspects of the token distribution.
By leveraging this rank-$r$ decomposition, we aim to enhance the accuracy of multi-token predictions, thereby increasing token acceptance rates during speculative decoding and reducing overall inference time.
Thus, our main contributions are as follows:
\begin{itemize}
    \item We identify the limitations of existing multi-token prediction methods that predict tokens independently.
    \item We introduce a novel model that employs a rank-$r$ canonical probability decomposition to better approximate the joint distribution of future tokens.
    \item We demonstrate that our approach can be integrated into existing transformer architectures with minimal overhead, resulting in more efficient sampling without significant increases in computational cost.
\end{itemize}





\begin{figure*}[t!]
    \centering
    \includegraphics[width=\textwidth]{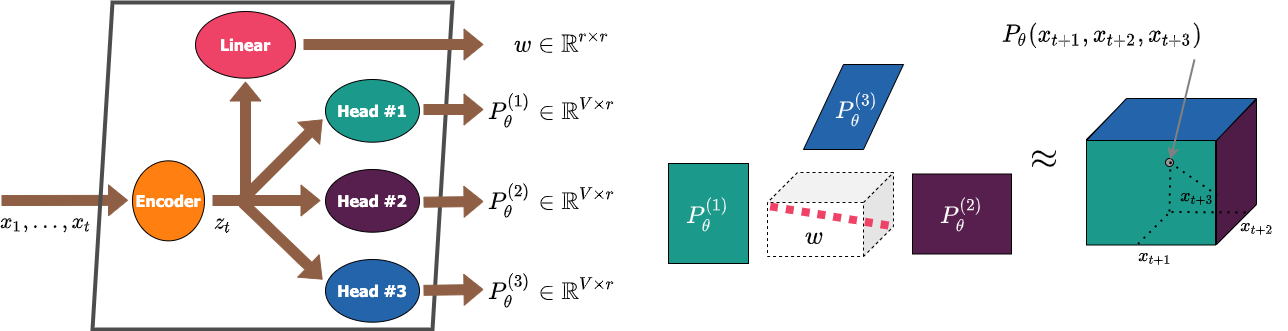}
    \caption{Schematic representation of the proposed model that predicts several tokens at once for a given sequence $x_{1}, x_{2}, \ldots, x_{t}$. We present the case of $n = 3$ predicted tokens $x_{t+1}, x_{t+2}, x_{t+3}$ and, accordingly, three heads which generate factor matrices $P_{\theta}^{(1)}$, $P_{\theta}^{(2)}$, and $P_{\theta}^{(3)}$ of the canonical decomposition and linear layer that generates weights $w$ are depicted.}
    \label{fig:scheme}
\end{figure*}
\section{Preliminaries}


In this section, we introduce the background and the formulation of the task we aim to solve through the approximation of a probability tensor using CP-decomposition.
For simplicity, this section assumes a greedy decoding approach; however, all presented methods are generalized to support probabilistic decoding.


Due to the autoregressive nature of sequence generation in large language models, the cost of generating a single new sample is approximately the same as the cost of generating multiple new samples that share a common prefix. 
This enables the default speculative decoding paradigm ~\citep{chen2023accelerating, leviathan2023fast}, which employs a Draft and then Verify mechanism to allow multiple tokens to be generated in one step, realizing lossless acceleration.

Given a model $M$, prefix $(x_1, x_2, \ldots, x_t)$ and a smaller draft model $M_d$ alligned with an original model, we can sequentially generate samples $(x^d_{t+1}, x^d_{t+2}, \ldots, x^d_{t+n})$ from $M_d$.
Then, we verify these samples by performing a single forward pass through the original model, generating the sequence $(x^c_{t+1}, x^c_{t+2}, \ldots, x^c_{t+n})$, where $x^c_{t+k} = M(x_1, x_2, \ldots, x_t, x^d_{t+1}, \ldots, x^d_{t_k-1})$.
The longest common prefix between $x^d_{t+1:t+n}$ and $x^c_{t+1:t+n}$ of the length $k$ becomes $x_{t+1}, \ldots, x_{t+k}$.
It is easy to see that these tokens are identical to the ones that would be generated by $M$ under the regular sampling procedure.
Additionally, the better the alignment between the smaller model $M_d$ and the original model $M$, the longer the common prefix, and thus the greater the speedup.



One possible extension of the described standard speculative decoding paradigm to increase the length of the accepted prefix is tree verification, as presented in~\cite{wang2024opt, xiong2024dyspec}.
In this approach, multiple sequences are generated from the draft model, and then all tokens are verified simultaneously using a tree-like custom attention mask.
The structure of the tree can either be static (with a fixed shape) or dynamic.



The draft model $M_d$ can also be obtained through modifications of the original model (self-speculative decoding) as in \cite{zhang2023draft, elhoushi2024layer, hooper2023speed}.
This bypasses the need to train a smaller model from scratch and opens up new possibilities for aligning base and draft model.

One way to obtain a draft model for a self-speculative decoding scenario is to modify the original model so that it predicts multiple tokens at once during training. This approach was implemented in a simple form in~\cite{gloeckle2024better}, and more recent works, like Deepseek v3~\cite{liu2024deepseek}, use a more sophisticated method for predicting multiple tokens.
From the perspective of the speculative decoding framework, however, the concept remains the same: we have a draft model $M_d$, which generates multiple tokens at once, and a target model $M$, where only the parameters responsible for generating the first token are activated.

In all of those cases we want our draft model $M_d$ to generate multiple tokens faster than the original model $M$, while ensuring that the generated tokens are as close as possible to the original ones.
Therefore, approximating the joint probability $P(x_{t+n:t+1} \vert x_{t:1})$ becomes an important challenge.
\section{Method}
    \subsection{Overall concept}
    We propose a model architecture that differs from traditional transformer models by enabling simultaneous prediction of multiple tokens through a rank-$r$ Canonical Polyadic (CP) tensor decomposition~\citep{harshman1970foundations} of the joint probability distribution.
In Figure~\ref{fig:scheme} we provide a corresponding schematic illustration, the content of which will be disclosed later in this section.

The joint probability of the next $n$ tokens given the input sequence $x_{t:1}$ can be represented as a $n$-dimensional tensor:
\begin{equation}
\begin{split}
A \in &
    \mathbb{R}^{V \times V \times \ldots \times V},
\\
A[x_{t+1}, \ldots, x_{t+n}] = &
    P_{\theta}(x_{t+n:t+1} \, \vert \, x_{t:1}),
\end{split}
\end{equation}
where $V$ is the vocabulary size.
The tensor $A$ encapsulates the probabilities of all possible combinations of the next $n$ tokens. 
In~\cite{gloeckle2024better} it was proposed to approximate this joint distribution by assuming that future tokens are conditionally independent given the past as shown in~\eqref{llmtelora:rank-1}.
We draw special attention to the fact that this may be interpreted as a rank-$1$ CP approximation to the tensor $A$.
While computationally efficient, such approximation ignores dependencies among the future tokens. 

To better capture these dependencies, we propose to approximate the joint distribution using a rank-$r$ CP tensor decomposition according to~\eqref{llmtelora:cp}.
In order to ensure that $P_{\theta}$ from this equation is indeed a probability tensor, it is sufficient to fulfill the normalization condition~\eqref{llmtelora:tens_norm}.
The difference between~\eqref{llmtelora:cp} and standard CP-decomposition is an additional constraint on the factors of decomposition, i.e., each factor, $P_{\theta}^{(s)}$ should be non-negative and sum up to $1$ along one mode:
\begin{equation}
\sum_{x_{t+s}=1}^V
    P_{\theta}^{(s)}(x_{t+s} \vert x_{t:1}, \alpha)
= 1,
\quad
s = 1, 2, \ldots, n.
\end{equation}
This is easily achieved by taking \emph{softmax} operation along the mode direction. 

Thus, for the given input sequence $x_{t:1}$ we compute its embeddings $e_{t:1}$ using the encoder of the autoregressive transformer model.
Focusing on the last embedding $e_t$, we aim to predict the next $n$ tokens by parametrizing the factors of the decomposition as simple functions of $e_t$.
We introduce $n$ heads each corresponding to one of the next $n$ tokens.
For each position $s=1, 2, \ldots, n$ the conditional probabilities are defined as:
\begin{equation}\label{llmtelora:model-arc}
P_{\theta}^{(s)}(x_{t+s} \vert x_{t:1}, \alpha) =
    \text{softmax}\left(
        W^{(s)}_{\alpha} e_t \right)_{x_{t+s}},
\end{equation}
where $W^{(s)}_{\alpha} \in \mathbb{R}^{V \times E}$ are the weight matrices for each head and component, $V$ is the vocabulary size and $E$ is the embedding dimension.
The mixture weights $w_{\alpha}$ are computed in a similar way using an additional linear layer:
\begin{equation}
w = \text{softmax}\left(
    W_h e_t \right),
\end{equation}
where $W_h \in \mathbb{R}^{r \times E}$.



\subsection{Training procedure}

In training, we maximize the log-likelihood of the predicted $n$ tokens.
The computation of the log-likelihood is straightforward: first, the embeddings are calculated by the transformer backbone (it has the same cost as for the next token prediction).
We need to evaluate the logarithm of the likelihood, so using~\eqref{llmtelora:model-arc} directly is not numerically stable.
Instead, we compute everything using the logarithms of the probabilities.
For each pair of sequences $x_{t:1}$ and $x_{t+n:t+1}$, we evaluate the logarithm of the mixture weights $w$ (the computational cost corresponds to a matrix-by-matrix product and logsoftmax operation), then use~\eqref{llmtelora:model-arc} to compute $n$ matrices of the size $V \times r$
\begin{equation}
C^{(s)}_{\theta, \alpha} = \log P_{\theta}^{(s)}(x_{t+s} \vert x_{t:1}, \alpha),
\end{equation}
to calculate logarithms of the conditional probabilities in a stable way with \emph{logsumexp} operation:
\begin{equation}\label{llmtelora:loss1}
\begin{split}
L
& =
\log \left(
    P_{\theta}(x_{t+n:t+1} \vert x_{t:1})
\right)
\\
& \approx
\log \left(
    \sum_{\alpha=1}^{r}
        w_{\alpha} \prod_{s=1}^{n}
        P_{\theta}^{(s)}(x_{t+s} \vert x_{t:1}, \alpha)
\right)
\\
& =
\log \left(
    \sum_{\alpha=1}^r
        w_{\alpha}
        \prod_{s=1}^n \exp(C^{(s)}_{\theta, \alpha})
\right)
\\
& =
\log \left(
    \sum_{\alpha=1}^r \exp \left(
        \log{w_{\alpha}} +
        \sum_{s=1}^n C^{(s)}_{\theta, \alpha}
    \right)
\right).
\end{split}
\end{equation}
\subsection{Auxilary load balancing loss}
    Each term of the summation in~\eqref{llmtelora:loss1} corresponds to a single \textbf{expert}, which predicts its own probabilities for each token.
We have found, that optimizing such loss directly leads to the effects, similar to the ones observed in Mixture Of Experts (MoE) framework~\citep{masoudnia2014mixture, cai2024survey}: one expert (i.e., rank-1 term in our case) dominates the others, leading to worser likelihood even in the presence of larger number of parameters.\footnote{
    We note, that such interpretation and connection is not well-known in the low-rank  approximation community, and can be investigated further on. 
}
To obtain the balance between different experts, we utilize the achievements from the MoE communities and propose to use an auxiliary balancing loss on $w$.

It is well known that a critical challenge in training MoE models is ensuring equitable utilization of all experts~\citep{zhou2022mixture}.
Without proper balancing, some experts may become dominant, handling a disproportionate share of the data, while others remain underutilized.
To address this, we incorporate an \textbf{auxiliary balancing loss}.
This auxiliary loss penalizes imbalances in the expert weights and encourages to distribute the workload evenly across all experts.

Formally, the auxiliary loss can be represented as:
\begin{equation}
\mathcal{L}_{\text{aux}} =
    \sum_{\alpha=1}^{r} \left(
        \frac{n_\alpha}{N} - \frac{1}{r}
    \right)^2,
\end{equation}
where: $r$ is the number of experts, $n_\alpha$ is the number of tokens with maximal weight on expert $\alpha$, and $N$ is the total number of tokens.
This formulation ensures that each expert $\alpha = 1, 2, \ldots, r$ is utilized approximately equally, mitigating the risk of certain experts becoming bottlenecks.

Empirical observations have demonstrated that training the model \textbf{without the auxiliary loss} or \textbf{using the auxiliary loss values proposed in previous works} leads to training instability and eventual failure.
The auxiliary loss is pivotal in maintaining a balanced distribution of token assignments among experts, which is essential for stable convergence and effective learning.
Therefore, careful tuning of the auxiliary loss coefficient is necessary to achieve optimal performance.
By ensuring balanced expert utilization through the auxiliary loss, the model enhances the accuracy of multi-token predictions, which increases token acceptance rates during speculative decoding, thereby reducing overall inference time.
\subsection{Sampling method}
    Our sampling scheme is similar to the one proposed in~\cite{gloeckle2024better}.
We sample candidates from the proposal distribution (our approximation to the joint distribution of the next tokens) and then accept them or reject according to the recommendations of the draft model (which is the same model that predicts the next token).

For the rank-$1$ case the sampling is easy: probability distributions are computed for each token independently, and sampling is done from the computed distributions.
For our canonical rank-$r$ representation we need to use sequential sampling which is autoregressive, but only works with the factors of decompositions.
This makes sampling $dim$ tokens from our rank $r$ model just a bit slower, than $1$ token from the base model.

Note that the first marginal distribution 
$P(x_{t+1} \vert x_{t:1})$ is given by the first head directly, and we just need to average among $\alpha$:
\begin{equation}\label{first head sampling}
P_{\theta}(x_{t+1}) = \sum_{\alpha=1}^r
    w_{\alpha}
    P_{\theta}^{(1)}(x_{t+1} \vert x_{t:1}, \alpha),
\end{equation}
which can be also computed using logsumexp operation.
From this distribution, we sample the first token $x_{t+1}$.

Given $x_{t+1}$ we can now compute the marginal distribution:
\begin{equation}
\begin{split}
P_{\theta}(x_{t+2} & \, \vert \, x_{t+1})
 = 
\\
\sum_{\alpha=1}^r
&
    w_{\alpha}
    P_{\theta}^{(1)}(x_{t+1} \vert x_{t:1}, \alpha)
    P_{\theta}^{(2)}(x_{t+2} \vert x_{t:1}, \alpha),
\end{split}
\end{equation}
which is also reduced to matrix-by-matrix products, \emph{logsoftmax} and \emph{logsumexp} operations, and can be implemented by updating the unnormalized logits of the experts with incorporation of $\log P_{\theta}^{(1)}(x_{t+1} \vert x_{t:1}, \alpha)$ into them.

The sampling of the following tokens is also straightforward.
Given sampled $x_{t+1}, \ldots, x_{t+s-1}$ we then compute the probability:
\begin{equation}
\begin{split}
P_{\theta}(x_{t+s} \, \vert \, & x_{t+1}, \ldots x_{t+s-1}) =
\\
& \sum_{\alpha=1}^r
    w_{\alpha}
    \prod_{k=1}^{s-1} P_{\theta}^{(k)}(x_{t+k} \vert x_{t:1}, \alpha).
\end{split}
\end{equation}

\subsection{Speculative decoding}
    One of the advantages of multi-token models is the ability to accelerate the decoding of the base model in speculative decoding paradigm without compromising accuracy (lossless acceleration). 
Our sampling method seamlessly integrates with the speculative decoding framework by enhancing its capacity to handle multi-token predictions.
The usual setup for a speculative decoding consists of a draft model and a base model.
In our case we implement a modification of a self-speculative decoding algorithm, as described in~\cite{zhang2023draft, elhoushi2024layer}.
So, as a base model we take a next-token prediction model and as a draft model -- the prediction for $dim$ tokens forward obtained from a full ``CP-head''. In some experiments we also use a simple static draft tree, where for each token we sample top-$n$ candidates. Any advanced tree building strategy such as~\cite{miao2024specinfer, xiong2024dyspec} can also be used.\footnote{
    Our work focuses on ``better draft model'' part of speculative decoding, not on the ``most effective token tree'' part, so we keep the token generation part simple.
}

Self speculative decoding with rank $r$ model inherits two nice small benefits from more simple rank-1 model: first generated sample from the draft model is always accepted and one additional token from the base model is generated.
This means, that in one pass of the draft model with the base model we will obtain at least 2 tokens.
Due to this fact it seems beneficial to use that type of models even with moderate quality of the draft model.

\subsection{Fine tuning}
    
In the case, when we already have a pretrained model $M$ and want to produce $M_d$ out of it by training only the new tensor head, we face a different challenges, as in the case, when entire model is trained from scratch. To address those challenges we propose two main modifications. 

\begin{figure}[htb]
    \includegraphics[width=\columnwidth]{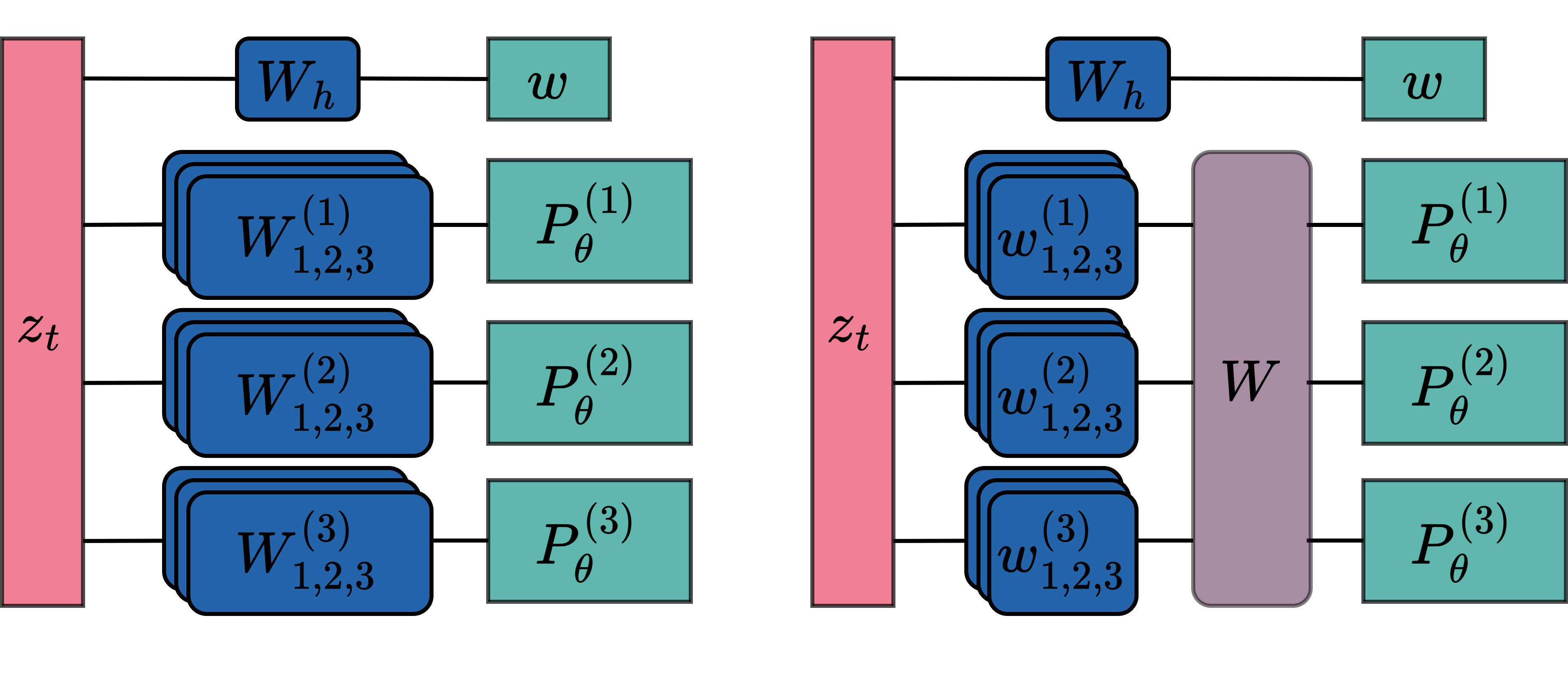}
    \caption{Comparison of original (left) and reduced (right) tensor head designs. $W$ is unchanged during fine-tuning process.}
    \label{fig:reduced_head}
\end{figure}

First one is the design of our custom tensor head. When we already have an initial head $W \in \mathbb{R}^{V \times E}$ then instead of taking $W^{(s)}_{\alpha} \in \mathbb{R}^{V \times E}$, we can use a shared head matrix $W$ a smaller $w^{(s)}_{\alpha} \in \mathbb{R}^{V \times E}$, and set $W^{(s)}_{\alpha} = W \times w^{(s)}_{\alpha}$. The difference can be seen in Figure~\ref{fig:reduced_head}. While this theoretically reduces the expressivity of our custom head, this approach is highly efficient, as it'll be demonstrated in the next section. Also, since the embedding size $E$ is usually at least an order of magnitude lower, than vocabulary size $V$, the total amount of parameters in tensor head is greatly reduced. 

The second one is the change in the training paradigm. Since now one have an access to the logits of the target model and those logits are unchanged during the training, one can use distillation loss in combination with targets given by an inital sequence.
Using the fact, that we want distribution, generated by our head be as close to initial head as possible, we write the distribution for the $k$-th token like 
$$
p^d_{t+k} \thicksim P_{\theta}^{k}(x_{t+k}|x_{1:t}, p^c_{t+1:t_k-1}),$$
where $p^c_{t+k} \thicksim P(x_{t+k} | x_{1:t+k-1})$ (distribution given by original head with respect to the previous tokens). Loss for the $k$-th token will look like 
\begin{equation}
L_k = \beta \cdot KL(p^d_{t+k}, p^c_{t+k}) + (1 - \beta) \cdot CE(p^d_{t+k}, x_{t+k}),
\end{equation}
where $\beta$ is a constant set to 0.9, $KL$ is a Kullback–Leibler divergence and $CE$ is a Cross-entropy. Then, the full loss looks like \\
$
     L = \sum_{i=0}^d \left( \gamma^i L_i \right) + \mathcal{L}_{\text{aux}},
$
where $\gamma$ is a discount parameter and $\mathcal{L}_{\text{aux}}$ is the same, as in case of training from scratch. This loss is inspired by the usual distillation logic and the loss used by \cite{gao2024falcon}.

\section{Experiments}
    In this section, we present a comprehensive evaluation of our proposed multi-token prediction approach.
Experiments are designed to assess the efficacy of different ranks and auxiliary loss configurations, the capability to fine-tune only the prediction head, and the impact on inference speed for large-scale models.

\subsection{Training different ranks and auxiliary loss models}

\begin{figure}[t!]
    \centering
    \begin{subfigure}[b]{0.49\textwidth} 
        \includegraphics[width=\textwidth]{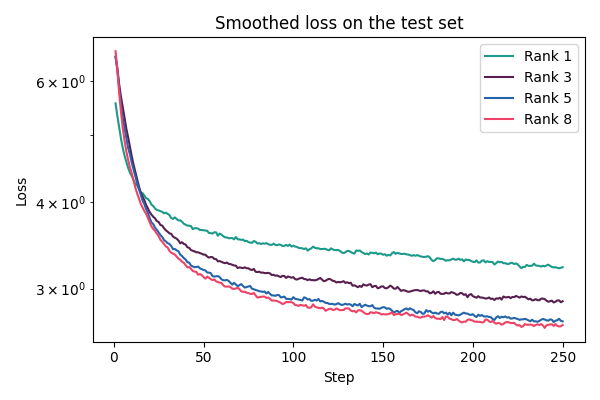} 
    \end{subfigure}
    \hfill
    \begin{subfigure}[b]{0.49\textwidth} 
        \includegraphics[width=\textwidth]{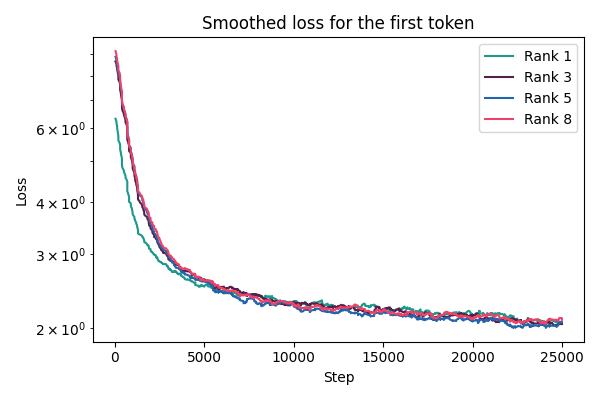} 
        
    \end{subfigure}
    \caption{Losses for the tiny transformer model with different CP-rank values trained on the TinyStories dataset.}
    \label{fig:tiny_stories_rank_dep}
\end{figure}

\begin{figure}[t!]
    \centering
    \begin{subfigure}[b]{0.49\textwidth} 
        \includegraphics[width=\textwidth]{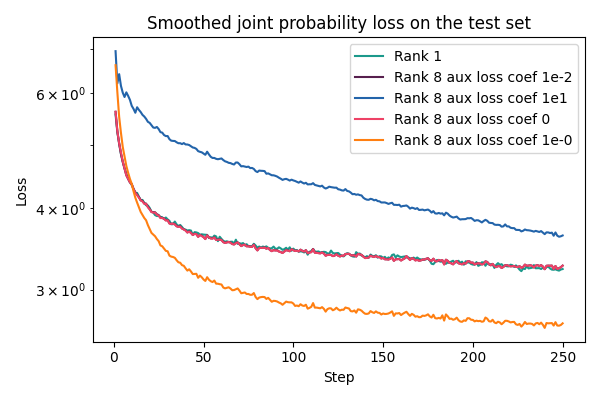} 
    \end{subfigure}
    \hfill
    \begin{subfigure}[b]{0.49\textwidth} 
        \includegraphics[width=\textwidth]{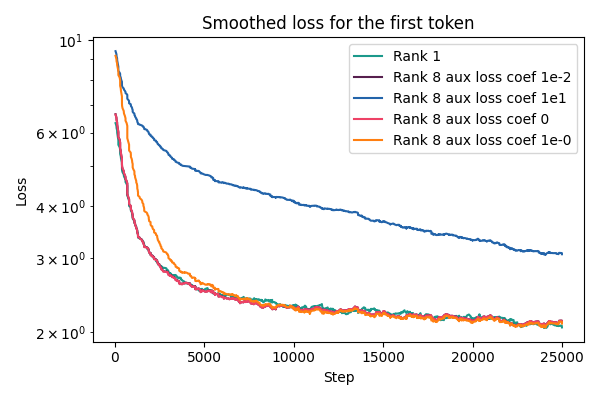} 
    \end{subfigure}
    \caption{Losses for the rank-8 tiny transformer model trained on the TinyStories dataset with different auxiliary loss penalties compared to the baseline (i.e., the rank-1 model).}
    \label{fig:tiny_stories_loss_scale_dep}
\end{figure}

For our experiments we've chosen the multi-head tiny transformer model with 56.3 M parameters based on the code in~\cite{Karpathy2022}.
We consider the case of $4$ heads and added RoPe positional encodings as in~\cite{su2024roformer}.
Training was conducted on the Tiny Stories dataset~\citep{eldan2023tinystories} using various ranks for the CP-decomposition.
The objective was to observe how increasing the rank influences the joint loss and loss on the first token. Because the quality of our final generation depends only on the quality of the first head, we tracked both those metrics.
Additionally, we experimented with different sizes of the auxiliary loss penalty to ensure balanced expert utilization.

As illustrated in the left graph in Figure~\ref{fig:tiny_stories_rank_dep}, increasing the rank from $1$ to higher values leads to a consistent decrease in joint loss, indicating a better approximation of the joint probability distribution.
This trend underscores the model's enhanced capability to capture inter-token dependencies with higher ranks.

Contrary to the joint loss, right graph in Figure~\ref{fig:tiny_stories_loss_scale_dep} shows that the loss for the first token remains largely unchanged across different ranks.
It is worth noting, that probability distribution for the first token as a function of last layer embeddings $m$ is given by $\sum_{\alpha=1}^{r} w_{\alpha}(m) C_{\alpha}(m)$ (in notation of~\eqref{llmtelora:loss1}) and both $C_{\alpha}$ and $w_{\alpha}$ are linear, which makes this function equivalent to a simple linear head.
So, after convergence we expected the same loss for all of the ranks.
As follows from the reported results, this is exactly what happened and this consistency confirms that our model maintains optimal training for the initial token prediction, ensuring that the foundational aspects of the sequence generation remain robust.
The loss on the first token is especially crucial, because with a speculative decoding we are improving sample for a big model, which is in our self-speculative case is a first head.
We also note that from Figure~\ref{fig:tiny_stories_rank_dep} it follows that all of the inference speedup will be obtained without compromising quality.

Figure~\ref{fig:tiny_stories_loss_scale_dep} presents the effect of varying the auxiliary loss penalty size. We observed that with a very small penalty, the joint loss mirrors that of the rank-1 model, suggesting insufficient balancing among experts. Conversely, an excessively large penalty led to prolonged convergence times, as depicted in the figure. Then we identified an optimal penalty size balancing expert utilization without hampering training accuracy.

\begin{figure}[t!]
    \centering
    \begin{subfigure}[b]{0.42\textwidth} 
        \includegraphics[width=\textwidth]{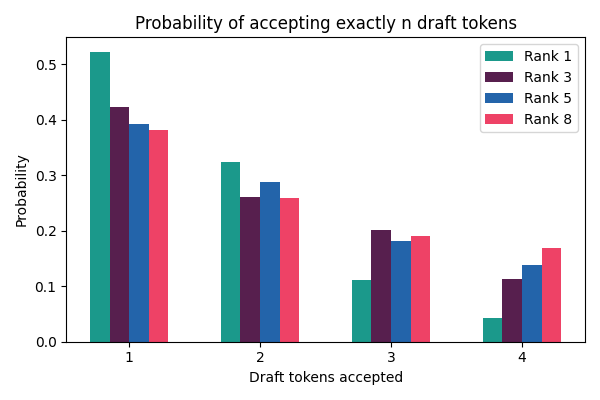} 
        \label{fig:left_image}
    \end{subfigure}
    \hfill
    \begin{subfigure}[b]{0.42\textwidth} 
        \includegraphics[width=\textwidth]{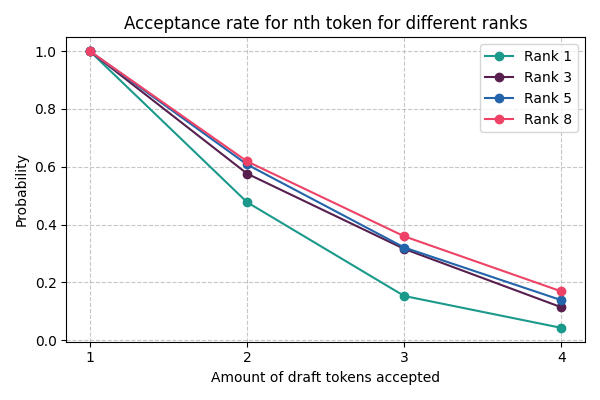} 
        \label{fig:right_image}
    \end{subfigure}
    \caption{Speculative decoding performance for the tiny transformer model with different CP-rank values trained on the TinyStories dataset from scratch.}
    \label{fig:side_by_side0}
\end{figure}

The efficiency of our model in speculative decoding was evaluated by measuring the acceptance rate of drafted tokens.
Figure~\ref{fig:side_by_side0} and Table~\ref{table:experiment_results} illustrates that on the Tiny Stories dataset, models with higher ranks achieved up to a around 30\% increase in accepted drafts.
This allowed us to reduce inference time even for this tiny (``nanoGPT'') model for which the head is responsible for a significant percentage of computational time, which is not the case for larger models.

To ensure scalability of our approach we've also conducted experiments on a bigger model. Namely, we took an 1B model, obtained by increasing the amount of hidden size and amount of transformer blocks of the model, implemented in \href{https://github.com/alexjc/nanogpt-speedrun} (GPT-2 architecture) and tuned the obtained model on the fineweb (\cite{penedo2024fineweb}) dataset. The results are presented in the Table~\ref{table:1B_model_results}. We've achieved 24\% increase in accepted draft length, when compared to a rank-1 model and 44\% reduction in execution time, when compared to a baseline.

\begin{table}[h]
    \centering
    \small
    \caption{Inference time comparison for 1B GPT2-like model}
    \label{table:1B_model_results}
    \begin{tabular}{lccp{2.5cm}}
    
        \hline
        \textbf{Model} & \textbf{Time (s)} & \textbf{Param. (B)} & \textbf{Avg. acceptance} \\
        \hline
        base & 43.8 & - & - \\
        rank 1 & 34.0 & 0.98 & 1.58 \\
        rank 2 & 28.9 & 1.2 & 1.96 \\
        rank 4 & 31.5 & 1.9 & 1.85 \\
        \hline
    \end{tabular}

\end{table}

\begin{table*}[t!]
\caption{Results with the speculative decoding for the tiny transformer model with different CP-rank values trained on the TinyStories dataset from scratch.}
\label{table:experiment_results}
\centering
\begin{tabular}{cccc}

\hline

\textbf{Rank} & \textbf{Loss} & \textbf{Avg. draft tokens accepted} & \textbf{Time per token (with speculative decoding)}\\
\hline

1 & 3.23 & 1.67 & 0.0336s \\

3 & 2.88 & 2.01 & 0.0328s \\

5 & 2.69 & 2.07 & \textbf{0.0303s} \\

8 & 2.66 & 2.15 & 0.0326s\\

\hline

\end{tabular}
\end{table*}

\subsection{Fine-tuning of pretrained model}

To evaluate the flexibility of our approach, we fine-tuned only the prediction head of the 1.82B model provided by the \cite{penedo2024the} using the $fineweb$-$edu$ dataset.
This experiment aimed to determine whether partial model updates could yield performance improvements without the computational overhead of full model fine-tuning.

\begin{figure}[t!]
    \centering
    \begin{subfigure}[b]{0.42\textwidth} 
        \includegraphics[width=\textwidth]{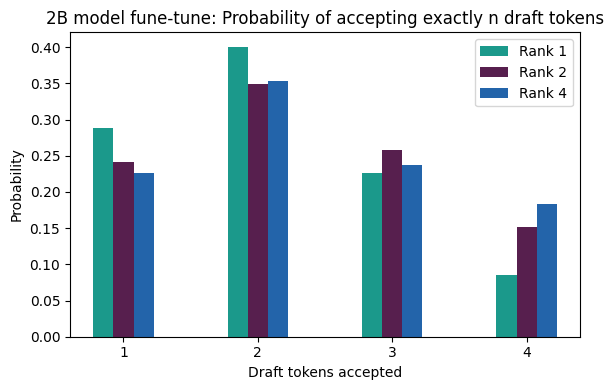} 
        \label{fig:left_image}
    \end{subfigure}
    \hfill
    \begin{subfigure}[b]{0.42\textwidth} 
        \includegraphics[width=\textwidth]{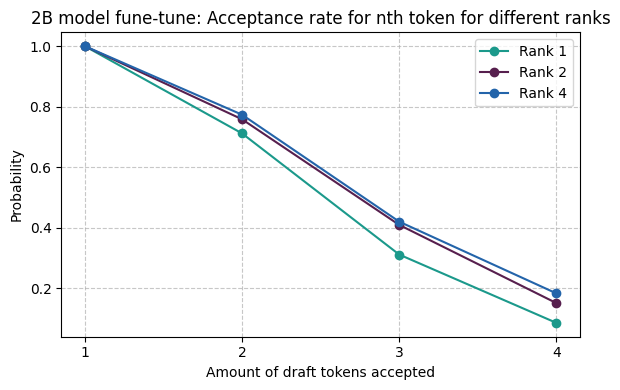} 
        \label{fig:right_image}
    \end{subfigure}
    \caption{Speculative decoding performance for fune-tuned head of 2B transformer model on fineweb-edu dataset with different CP-rank values.}
    \label{fig:side_by_side}
\end{figure}

Figure~\ref{fig:side_by_side} and Table~\ref{table:pycode_avg_draft_tokens} demonstrates speculative decoding performance for the experiments we conducted for different rank values. The results were obtained on temperature 0 using a simple uniform draft tree with 5 branches.
From the reported results it follows that even when only the head is fine-tuned, increasing the rank leads to marginal improvements in joint loss.
Additionally, we can see that speculative decoding benefits from higher ranks, albeit to a lesser extent (approximately 13\% increase in accepted drafts) compared to the full model training.

\subsection{Inference time benchmarking}

\begin{table}[t!]
\caption{Average number of accepted draft tokens and benchmark time for fune-tuned model.}
\label{table:pycode_avg_draft_tokens}
\centering
\begin{tabular}{ccc}

\hline

\textbf{Rank} & \textbf{Execution time (s)} &\textbf{Avg. Draft Tokens} \\

\hline

1 & 39.3 & 2.10 \\

2 & 36.9 & 2.32 \\

4 & 36.2 & 2.38 \\

\hline
\end{tabular}
\end{table}

\begin{table*}[t!]
\caption{Inference time for one forward pass comparison for Llama and Rocket models.}
\label{table:inference_time_comparison}
\centering
\begin{tabular}{cccccc}

\hline

\textbf{Rank} 
& \textbf{Llama 8B Barebone}
& \textbf{Llama 8B Head}
& \textbf{Llama 8B Full} & \textbf{Rocket 3B Full} \\

\hline

Barebone & 0.1761 & - & 0.1761s & 0.0154 \\

Rank 1 & 0.1761 & 0.0132 & 0.1893 & 0.0160 \\

Rank 3 & 0.1825 & 0.0129 & 0.1954 & 0.0162 \\

Rank 5 & 0.1865 & 0.0330 &  0.2195 & 0.0166 \\

\hline

\end{tabular}
\end{table*}

To determine the impact of modified head on the inference time of bigger language models we benchmarked the time of one forward pass of our approach on large-scale models with 3 billion and 8 billion parameters. 
As reported in Table~\ref{table:inference_time_comparison}, the inference overhead for integrating the proposed multi-head layer remains minimal, even as the rank increases.
Note that for a CP-rank value of 5, we observe a significant increase in head execution time when considering the Llama model, likely due to its large vocabulary size. However, for moderately sized networks, inference time remains manageable and increases only slightly with higher CP-rank values. The obtained results align with the theoretical algorithmic complexity of our new layer. During inference, the computational complexity of the barebone model grows linearly (given KV caches), whereas the computational complexity of a rank-$r$ head remains constant. Our measurements were conducted with sequence lengths varying from 1024 to 4096, but in many practical applications, sequence lengths are even longer, further justifying the use of rank-$r$ heads in models with large context windows.
\section{Related work}
    Training large language models (LLMs) to predict multiple tokens all at once and in parallel can drive these models toward better sample efficiency.
Various approaches for multi-token predictions have been proposed recently.
In~\cite{stern2018blockwise} several feed-forward decoder layers from the last encoder state are added for prediction of the next several tokens simultaneously, and in~\cite{miao2024specinfer} this idea was further improved within the framework of the so-called Medusa heads that use tree attention mechanism.
In a number of works~\cite{song2021accelerating, santilli2023accelerating, fu2024break} it is proposed to generate multiple draft tokens in parallel on the basis of the Jacobi iteration methods, i.e., via solving a non-linear system of equations while auto-regressive decoding in LLM.
In the work~\cite{bhendawade2024speculative} multiple tokens are predicted by adding streaming embeddings initialized from upper layers, with the token tree reduced by early exiting.

Thus, this direction of research is actively developing today, however, the approaches outlined above have several limitations, including the need for significant changes in the original architecture of the model and limited speedup.
Therefore, of particular interest is the recent work~\cite{gloeckle2024better}, where it was proposed to approximate the joint probability of the next several tokens using multiple heads but assuming conditional independence given the previous context.
As we have already noted above, this approach remains a crude approximation that may limit token acceptance rates in the speculative decoding approach due to its disregard for token interdependencies.
To improve upon this, we considered in this work a more accurate approximation of the joint distribution in the form of the CP-decomposition.

To effectively implement the proposed scheme, we paid attention to connection of the used weighted CP-decomposition with the Mixture of Experts (MoE) technique. MoE is a widespread approach to enhance capabilities of LLMs with the most popular one being Sparse-Gated MoE introduced in~\cite{shazeer2017outrageously}.
MoE implementations can be either sparse or dense with sparse version being more popular, but there are many usages of both options, as in~\cite{dou2023art} and~\cite{pan2024dense}.
While many parameters of MoE approach can be tweaked~\citep{cai2024survey}, the most common option is using MoE inside a transformer block, as in~\cite{zhou2022mixture}.

In this work, as an application of the proposed model for multi-token prediction, we consider its use as part of the speculative decoding scheme, which was proposed in~\cite{leviathan2023fast} and nowadays has become a common technique in the domain of inference acceleration.
Self speculative decoding and multi-token prediction naturally go well with each other.
This combination may require modification in model architecture as in~\cite{bhendawade2024speculative}, but it is possible to modify only heads as in~\cite{gloeckle2024better, li2024eagle} to enable faster application of the approach to existing LLMs, and we use such approach. 
\section{Conclusion}
    In this work, we propose a new model for multi-token prediction in transformers based on the Canonical Polyadic (CP) tensor decomposition of the joint probability distribution. 
The results indicate that our model can be efficiently trained across a wide range of ranks, with higher ranks consistently yielding lower joint losses. This improvement underscores the model's ability to better capture the dependencies among future tokens, leading to more accurate predictions.

We observed a direct correlation between lower joint losses and enhanced speculative decoding performance.
Specifically, our approach significantly increased the acceptance rates of predicted tokens, with notable improvements of up to 50 \%  of draft tokens accepted.
The factor matrices of our decomposition of the joint probability tensor are generated by several heads that use shared model trunk, which practically makes it possible to minimize extra costs during inference and convert higher draft token acceptance to faster inference times.

The ability to fine-tune only the prediction head of the model while maintaining competitive performance highlights the flexibility of our approach.
This capability allows for targeted improvements without the computational overhead associated with full model retraining.
Benchmarking inference speed for bigger models demonstrated that our method introduces negligible inference overhead, ensuring that in many practical cases the benefits of improved performance for draft model do not come at the cost of increased latency.  
\bibliography{biblio}
\bibliographystyle{icml2025}
\end{document}